\documentclass{article}
\usepackage[usenames,dvipsnames,svgnames,table]{xcolor}
\usepackage[breaklinks=true,letterpaper=true,colorlinks,bookmarks=false,citecolor=MidnightBlue,linkcolor=BrickRed]{hyperref}
\usepackage{nips13submit_e,times}
\usepackage[round]{natbib}
\usepackage{bibhacks}
\usepackage{graphicx}
\usepackage{subfigure}
\usepackage{amsmath}

\usepackage[compact]{titlesec}
\titlespacing{\section}{0pt}{0ex}{0ex}
\newcommand{\vs}{\emph{vs.}}

\DeclareUrlCommand\emailurl{}
\newcommand{\email}[1]{\href{mailto:#1}{\emailurl{#1}}}

\title{Felzenszwalb-Baum-Welch: Event Detection through Changing Appearance}

\author{
Daniel Paul Barrett\thanks{School of Electrical and Computer
  Engineering, Purdue University, West Lafayette IN 47907-2035}\\
\email{dpbarret@purdue.edu}\\
\And
Jeffrey Mark Siskind\footnotemark[1]\\
\email{qobi@purdue.edu}
}

\nipsfinalcopy 

\begin{document}

\maketitle

\begin{abstract}
We propose a method which can detect events in videos by modeling the change
in appearance of the event participants over time.
This method makes it possible to detect events which are characterized not by
motion, but by the changing state of the people or objects involved.
This is accomplished by using object detectors as output models for the states
of a hidden Markov model (HMM).
The method allows an HMM to model the sequence of poses of the event
participants over time, and is effective for poses of humans and inanimate
objects.
The ability to use existing object-detection methods as part of an event model
makes it possible to leverage ongoing work in the object-detection community.
A novel training method uses an EM loop to simultaneously learn the temporal
structure and object models automatically, without the need to specify either
the individual poses to be modeled or the frames in which they occur.
The E-step estimates the latent assignment of video frames to HMM states, while
the M-step estimates both the HMM transition probabilities and state output
models, including the object detectors, which are trained on the weighted
subset of frames assigned to their state.
A new dataset was gathered because little work has been done on events
characterized by changing object pose, and suitable datasets are not
available.
Our method produced results superior to that of comparison systems on this
dataset.
\end{abstract}

\section{Introduction}

The problem of event detection is identifying instances of particular actions
in video.
Many every-day events are defined as state changes.
For example, `to open' is to cause a state change between an object being
closed and being open.
Opening a drawer involves a state change from a drawer contained within a
desk to a drawer sticking out from the desk.
Much work in event detection focuses on finding characteristic spatio-temporal
patterns, to the point where that has even been stated as the goal
\citep{ke2007event}
However, there is no characteristic motion pattern for an event as seemingly
simple as `open'.

We have modeled events by the characteristic pose sequences of the event
participants.
An appearance model is automatically learned for each of the characteristic
poses of an object during an event.
Thus, we detect the opening of a drawer by detecting the transition of the
drawer from matching a closed drawer model to matching an open drawer model.

There are at least two major categories of event detection and classification
systems: object-based, and statistical-local-feature-based approaches.
Statistical-local-feature-based approaches attempt to apply a classifier
directly on low-level features, such as STIP \citep{willems2008efficient},
extracted from the video.
These approaches cannot distinguish the features extracted from objects from
those from the background, and often detect events for artifactual reasons,
such as detecting events like `dive' by a blue background
\citep{liu2009recognizing}.
They also cannot distinguish between the features of multiple objects, and so
are not well suited to detect multiple simultaneous events.
Further, these methods tend to rely on motion features.

Object-based approaches attempt to first detect and track the event
participants.
They then extract features from those tracks, and then apply a classifier to
those features.
The use of tracks makes it possible to do more than
obtain a simple event label, as shown in \citet{Barbu2012b}, where entire
sentences, including subject, object, adjectives, adverbs, and prepositional
phrases are automatically produced from video.
Some tracking-based approaches use the absolute or relative motion or position
of the object tracks or the optical flow in the region of the object as
features \citep{Barbu2012a,Barbu2012b}.
Such features work well for events such as one person giving an object to
another person or picking up an object.
However, for events such as bending over, opening a drawer or tipping over a
cup, that aren't characterized by gross object motion, such features are not
sufficient.

There is work in extracting human pose \citep{Yang2011}.
However, these methods are not generally used for event detection, because the
output is too noisy.
Using a single general pose model to detect many events requires that model to
be able to handle many different poses.
In contrast, our method automatically learns specific pose models for each
characteristic pose for each event.
Thus each model is well suited to detect its associated pose.

Our method is an object-based method, first detecting and tracking the event
participants.
We use hidden Markov models (HMM) to model the features extracted from the
tracks.
HMMs are commonly used for event detection
\citep{tang2012learning,Barbu2012a,Barbu2012b,Wang2009a,Xu2005,ren2009state}.
The main features that distinguish our approach from previous HMM-based methods
are:

(1) The inclusion of an object detector as the output model for a state in an
HMM, which makes it possible to model the sequence of object poses
characteristic of an event.

(2) A novel training method wherein the object models are learned as part of
the HMM state output model update in an EM loop.

Our method allows both the HMM parameters and object models to be learned,
without knowing beforehand either what poses are to be learned or in which
frames of the training videos they occur.
The object detectors are trained on a weighted selection of images as
determined by the current estimate of the latent state sequence over the course
of the EM loop.
As the loop iterates, the latent state sequence and object models are
reestimated, resulting in the object detector for each HMM state becoming
associated with one of the distinctive poses of the event participant.

Our use of object tracks makes it possible to detect multiple simultaneous
events, and to avoid classifying based on artifactual background features.
In contrast to previous object-based event detection, we do not rely on motion
features, but instead learn appearance models for each event state, making it
possible to detect events characterized by object state transitions, rather
than motion.
High-level event information is used to guide the training of these low-level
appearance models by training each specifically to model the pose seen in the
frames which are likely to be associated with the particular corresponding HMM
latent state.
A set of simple models trained to detect the specific poses relevant to a
particular event can work better for detecting those poses than attempting to
use a general human pose model.
In addition, our method works for objects other than people with no
modification.

\section{Theory}

\subsection{Brief Overview of Hidden Markov Models}

The hidden Markov model is a generative probabilistic model consisting of $n$
states $q_{1},\ldots,q_{n}$ which defines a stochastic process whereby a
sequence of $T$ observations $O_{1},\ldots,O_{T}$ can be generated.
Each state $q_{i}$ has an output distribution $b_{i}(O)$ and a transition
distribution $a_{ij}$.
$b_{i}(O)$ defines the probability of state $q_{i}$ generating observation
$O$.
$a_{ij}$ defines the probability of the model transitioning from state $q_{i}$
to $q_{j}$.
The parameters of the transition and output models are together denoted
$\lambda$.
The sequence of latent variables $X_t$ denote the state of the HMM at time $t$.

The likelihood of the observations $P(O_{1},\ldots,O_{T}|\lambda)$ is used to
determine whether they correspond to the event modeled by the HMM.
It and the distribution of the latent state sequence
$\gamma_{t}(i)=P(X_{t}=q_{i} | O_{1},\ldots,O_{T},\lambda)$ can be computed
efficiently with the Forward-Backward algorithm
\citep{Baum1966,Baum72,BaumPSW70}.

The solution is written in terms of two variables
$\alpha_{t}(i)=P(O_{1},\ldots,O_{t},X_{t}=q_{i}|\lambda)$, and
$\beta_{t}(i)=P(O_{t},\ldots,O_{T} | X_{t}=q_{i},\lambda)$, which can be computed
efficiently.

\begin{eqnarray*}
  \alpha_{t}(j)&=&\sum_{i=1}^{N}\alpha_{t-1}a_{ij}b_{j}(O_{t})\\
  \beta_{t}(i)&=&\sum_{j=1}^{N}a_{ij}b_{j}(O_{t+1})\beta_{t+1}(j)\\
  \gamma_{t}(i)&=&\frac{\alpha_{t}(i)\beta_{t}(i)}{\displaystyle
    \sum_{j=1}^{N}\alpha_{T}(i)}\\
  P(O_{1},\ldots,O_{T}|\lambda)&=&\sum_{i=1}^{N}\alpha_{T}(i)
\end{eqnarray*}

Learning an HMM which maximizes the output likelihood of a set of observation
sequences is done using the Baum-Welch
\citep{Baum1966,Baum72,BaumPSW70} algorithm, an iterative algorithm which
alternates between two steps, labeled E and M.

In the E step, the latent probabilities are computed as in the above equations.
In the M step, the HMM parameters are updated to increase the likelihood given
the latent variables.
The HMM transition probabilities are computed via:
\begin{eqnarray*}
  \xi_{t}(i,j)&=&\frac{\alpha_{t}(i)a_{ij}\beta_{t+1}(j)b_{j}(O_{t+1})}
     {\displaystyle\sum_{i=1}^{N}\sum_{j=1}^{N}
       \alpha_{t}(i)a_{ij}\beta_{t+1}(j)b_{j}(O_{t+1})}\\
     a_{ij}&=&\frac{\displaystyle\sum_{t=1}^{T}\xi_{ij}(t)}
     {\displaystyle \sum_{j=1}^{N}\sum_{t=1}^{T}\xi_{ij}(t)}
\end{eqnarray*}
The HMM output distributions $b_{i}(O)$ are also updated.
The specifics depend on the output model assumed for the HMM states.
Typically, the observations in an HMM are vectors of discrete or real numbers
produced by some method(s) of feature extraction, such as the velocity of an
object in the video.
The state output models $b_{i}(O)$ are vectors of either discrete or continuous
probability distributions over these features.
If each distribution in the vector $b_{i}(O)$ is updated in such a way that the
output likelihood given the latent variables of the observations of its feature
increases, then the EM loop is guaranteed to increase the total observation
likelihood.

\subsection{Object Detector as a State Output Model}

The image inside the bounding box of an object track in a video
frame can be used as one of the observed features in $O$.
The probability of this image being generated by an HMM state can be modeled
with an object-detector model corresponding to that state.
If an image is generated by an HMM state with probability proportional to the
degree to which it matches that state's object model, then the output
probabilities $b_{i}(O_t)$, can be obtained by computing the matching score
between the object model for each state $i$ and each image $O_t$.
To be treated as a probability, the score should be normalized between zero and
one.

This model allows the object detector to be updated in the EM loop by:
\begin{equation}
\label{traindpm}
  b_{i}(O)=\textsc{UpdateObjectDetector}(O,\gamma(i))
\end{equation}
where $\textsc{UpdateObjectDetector}(O,\gamma(i))$ is the method by which a
particular object model is learned from a set of weighted training images.
The weight of each sample $O_{t}$ passed into the training procedure for the
object detector of state $i$ is the probability $\gamma_{t}(i)$ that the HMM is
in state $i$ during that frame $t$ of the video.
As with standard HMM updates, if this function increases the output likelihood
of the images, then the total observation likelihood will increase in each step
of the EM loop.

\begin{figure}
  \begin{center}
    \begin{tabular}{@{}ccccc@{}}
      \includegraphics[scale=.3]{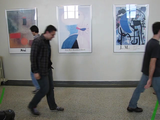}&
      \includegraphics[scale=.3]{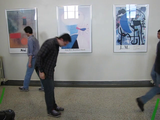}&
      \includegraphics[scale=.3]{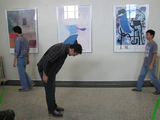}&
      \includegraphics[scale=.3]{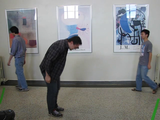}&
      \includegraphics[scale=.3]{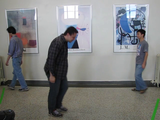}\\[-0.2ex]
      \includegraphics[scale=.6]{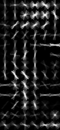}&
      \includegraphics[scale=.6]{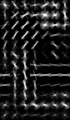}&
      \includegraphics[scale=.6]{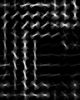}&
      \includegraphics[scale=.6]{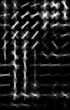}&
      \includegraphics[scale=.6]{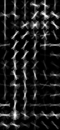}\\
      \hline\\[-1.2ex]
      \includegraphics[scale=.3]{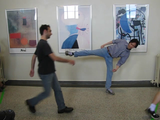}&
      \includegraphics[scale=.3]{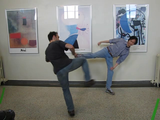}&
      \includegraphics[scale=.3]{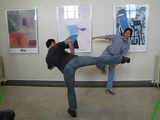}&
      \includegraphics[scale=.3]{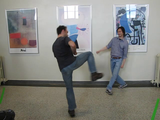}&
      \includegraphics[scale=.3]{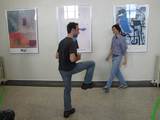}\\[-0.2ex]
      \includegraphics[scale=.6]{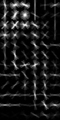}&
      \includegraphics[scale=.6]{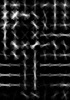}&
      \includegraphics[scale=.6]{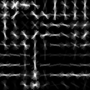}&
      \includegraphics[scale=.6]{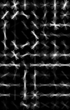}&
      \includegraphics[scale=.6]{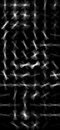}\\
      \hline\\[-1.2ex]
      \includegraphics[scale=.3]{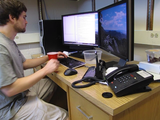}&
      \includegraphics[scale=.3]{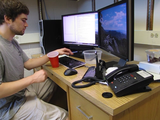}&
      \includegraphics[scale=.3]{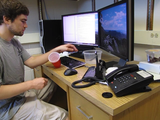}&
      \includegraphics[scale=.3]{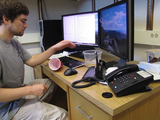}&
      \includegraphics[scale=.3]{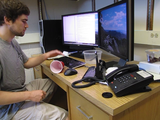}\\[-0.2ex]
      \includegraphics[scale=.6]{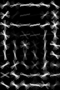}&
      \includegraphics[scale=.6]{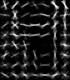}&
      \includegraphics[scale=.6]{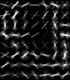}&
      \includegraphics[scale=.6]{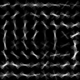}&
      \includegraphics[scale=.6]{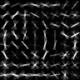}\\
      \hline\\[-1.2ex]
      \includegraphics[scale=.3]{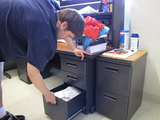}&
      \includegraphics[scale=.3]{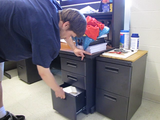}&
      \includegraphics[scale=.3]{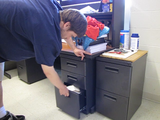}&
      \includegraphics[scale=.3]{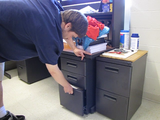}&
      \includegraphics[scale=.3]{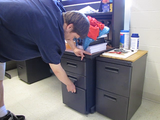}\\[-0.2ex]
      \includegraphics[scale=.6]{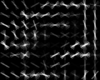}&
      \includegraphics[scale=.6]{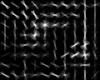}&
      \includegraphics[scale=.6]{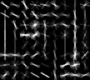}&
      \includegraphics[scale=.6]{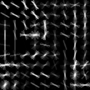}&
      \includegraphics[scale=.6]{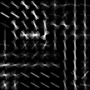}\\
    \end{tabular}
  \end{center}
  \caption{Example frames from training videos and visualizations of the object
    models for several 5-state HMMs.
    The first row in each section contains relevant frames from a training
    example and the second is a visualization of the object models from
    corresponding 5-state HMMs.}
  \label{fig:model-visualizations}
\end{figure}

In each step, the assignment $\gamma_t(i)$ of video frames to each state is
updated based upon the current HMM parameters.
The object models for each state are then updated based upon their own set of
weighted frames.
The sets of frames are differentiated and thereby the object models become
differentiated as the loop iterates.
The differentiation of models causes further differentiation in the frame
assignment, gradually resulting in each state's object model corresponding to a
particular pose which occurs during the event.
Visualizations of several resulting models can be seen in
\ref{fig:model-visualizations}.

\section{Implementation}

In order to illustrate the power of the novel contribution the proposed
method, we have not used any features other than the pose modeled by the object
detectors in our HMMs.

\subsection{Object Detector}

Any object detector which returns scores in a range between zero and one and
which can be trained using weighted samples can in principle be used with our
method.
The Deformable Parts Model (DPM;\citealp{Felzenszwalb2010b,Felzenszwalb2010a})
has recently been prominent in the object detection community.
It is largely scale invariant, due to is use of a scale pyramid, is
invariant to horizontal reflection, and is able to handle in-class variation
with the use of deformable parts.
Being based on Histogram of Oriented Gradients (HOG) descriptors
\citep{dalal2005histograms}, it is largely lighting and color invariant.
For these reasons, we have chosen to use DPM models in our HMMs.
Therefore our \textsc{UpdateObjectDetector}$(O,\gamma(i))$ update is the DPM
training procedure.

DPM actually meets neither of our stated criteria: its training procedure
neither accepts weighted training data, nor does the detector return scores
between zero and one.
Significant modification to the original approach of
\citet{Felzenszwalb2010b,Felzenszwalb2010a} was necessary to meet these
requirements.

The first criteria was achieved by normalizing the matching scores to $(0,1]$
based on the range of values seen in the training data.

The second criteria was achieved by modification to the
\citet{Felzenszwalb2010b,Felzenszwalb2010a} training procedure, which is an
SVM-based method.
It takes as input a set of positive and a set of negative training samples.
Each sample is a rectangular subset of an image.
Our positive samples are the images $O$ within the bounding boxes of object
tracks in the training videos, while the negative samples are obtained randomly
from the training videos.

Ignoring some subtleties, the DPM training method uses the standard SVM elbow
loss function:
\begin{equation*}
L=\sum_{i=1}^{N}\max(0,1-w\cdot x_{i})
\end{equation*}
where $N$ is the number of training samples, $w$ is the weight vector being
trained, and $x_{i}$ are the feature vectors of HOG descriptors computed from
the training samples multiplied by 1 or -1 for positive and negative samples,
respectively.
This cost function does not support weighted training data, as is necessary
for inclusion in the EM loop.
Therefore it was modified to do so.
The loss for a given sample is multiplied by its weight $\gamma_{i}$.
\begin{equation*}
L=\sum_{i=1}^{N}\gamma_{i}\max(0,1-w\cdot x_{i})
\end{equation*}

This allows the M step of the EM loop to train different models for each of
the states given the same set of images, according to the weights $\gamma_{i}$.

The EM loop is a maximum likelihood approach.
However, the DPM training function is discriminative.
When the models are updated, the difference in the models' scores
between the positive and negative samples is increased.
Therefore, using this model causes the method to be a hybrid of the
discriminative and maximum likelihood approaches and does not guarantee
that the EM loop will increase the likelihood of the model in each iteration.

\subsection{HMM Topology}

The HMM topology was limited to bi-diagonal transition matrices.
An additional end state was enforced after the end of the observation sequence.
These two constraints together force the HMM to go through the entire sequence
of states, resulting in low likelihood for sets of frames depicting a subset of
an event.

\subsection{Inputs and Outputs}

The input to the training system for a single event model is a set of videos
labeled with bounding boxes of object tracks and the start and end frames of
instances of that event class.
The output is an HMM transition matrix, DPM models for each of its states,
and a likelihood threshold used for event detection.

The input to the event detection system is the output of the training system,
an object tracker, and a video.
The output is a set of tuples each consisting of an event label, a start frame,
an end frame, and a set of bounding boxes of the track(s) in which the event
was detected.

\subsection{Event Detection}

The first step is to obtain object tracks, which each consist of a set of
bounding boxes each denoting the location of an object in a frame of the video.
Tracking of multiple objects in cluttered environments under occlusion is a
difficult and unsolved problem.
We have used the detection-based object tracker of
\citet{Barbu2012a,Barbu2012b}.
We also used hand annotated tracks to compare the performance of the
proposed method with perfect tracks from the performance under the influence
tracking errors.

The second step is to obtain the output probabilities $b_{j}(O)$ for each state
of each event model.
This is done by obtaining normalized object-detector scores for each bounding
box of each track from the models associated with each state of each event
model.

Next, the HMM likelihood function is computed for a set of time intervals
along each track.
The starting position and length of the intervals are both varied within a
fixed range.
The starting position was quantized to 10 frames and varied across the entire
video.
Interval lengths were varied between 10 and 50 frames, also subject to 10 frame
quantization.

Once HMM likelihoods have been obtained for a given model, non-maximum
suppression is applied.
Each interval corresponds to a 3 dimensional space-time volume.
The overlap measure used between two such volumes is the intersection of their
volumes divided by their union.
An overlap threshold of $0.5^{3/2}$ in 3D is analogous to the standard 2D
PASCAL VOC \citep{Everingham10} overlap threshold of .5 and is the threshold
used for this non-maximum suppression.

Finally, the likelihoods of those intervals which remain are compared against
the model's trained threshold.
This threshold was learned by maximizing the F1 score on the training videos.

\begin{figure}
\centering
  \mbox{
\subfigure[Open]{

    \begin{tabular}{ccc}
      \includegraphics[width=0.1\textwidth]{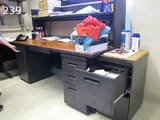}&
      \includegraphics[width=0.1\textwidth]{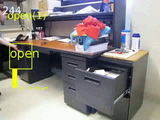}&
      \includegraphics[width=0.1\textwidth]{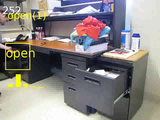}\\
      \includegraphics[width=0.1\textwidth]{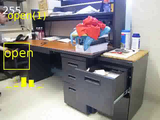}&
      \includegraphics[width=0.1\textwidth]{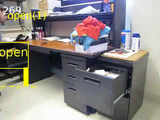}&
      \includegraphics[width=0.1\textwidth]{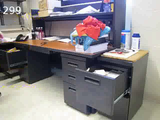}\\
    \end{tabular}
    \label{fig:detection-example1}
}\quad

\subfigure[Kick]{

    \begin{tabular}{ccc}
      \includegraphics[width=0.1\textwidth]{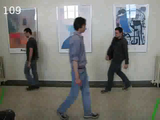}&
      \includegraphics[width=0.1\textwidth]{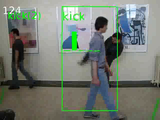}&
      \includegraphics[width=0.1\textwidth]{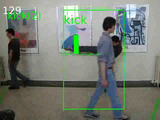}\\
      \includegraphics[width=0.1\textwidth]{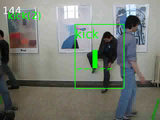}&
      \includegraphics[width=0.1\textwidth]{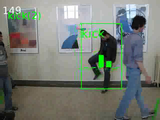}&
      \includegraphics[width=0.1\textwidth]{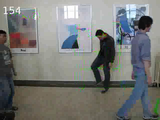}\\
    \end{tabular}
    \label{fig:detection-example2}
}}
\caption{Several frames illustrating detections of an open and
  a kick using automatic tracks}
\label{fig:detection-example}
\end{figure}

Those intervals whose likelihood does not surpass the likelihood threshold are
discarded.
Those remaining are given an event label.
These labeled intervals can be compared against human annotations to obtain
quantitative performance metrics, as will later be shown, or rendered with
histograms of the state probabilities $\gamma_{t}(i)$ on the video to
illustrate the output.
Fig.~\ref{fig:detection-example1} and~\ref{fig:detection-example2} show
automatic renderings of several frames from two examples of detected events.

\section{Experiment}

\subsection{Dataset}

There are two kinds of existing datasets for event detection.
First, there are those for which pose change is relevant, such as KTH
\citep{Schuldt2004} and Ballet \citep{Wang2009b}, but on which state of the art
systems, such as Action Bank \citep{SaCoCVPR2012} and C2 \citep{Kuehne11},
against which we compare our system, achieve near perfect performance.
Because of this, it would be impossible to show that our method outperforms
these systems by reporting results on these datasets.
The other kind of dataset is difficult, like HMDB \citep{Kuehne11} and Youtube
\citep{Liu2009}, but pose does not play a crucial role in identifying the
actions they represent.
As such they are unsuitable for demonstrating the efficacy of our approach,
which has been intentionally stripped of all features other than pose.

We therefore collected a dataset which is both difficult and for which pose
plays a crucial role.
It is difficult because it involves long videos consisting of large numbers of
event instances (about 30 per video) which often are simultaneous and/or are
partially occluded.
It also contains a reasonably large number of event event classes (11).

Three video scenes were filmed for both training and testing.
The first is a hallway scene depicting a number of people simultaneously
walking back and forth while periodically performing one of four actions:
bending, lunging, kicking, and waving.
The second scene depicts a cluttered office environment including seven
drawers of different shapes and at different scales.
The drawers were repeatedly opened and closed in a haphazard order by a
person.
The last scene depicts a person sitting at a cluttered desk, who operates a
hole punch, answers a phone, talks on a phone, hangs up a phone, and
knocks a cup over.
All scenes include significant clutter and occlusion of objects and people.

Two long videos of each scene were filmed: one for training and one for
testing, each containing many instances of each event.
The number of event instances can be seen in table \ref{table:dataset}

\begin{small}
\begin{table}
\begin{center}
\begin{tabular}{lr | lr | lr}
Event & \#Instances & Event & \#Instances & Event & \#Instances \\ \hline

Bend & 17 & Open & 32 & Answer & 10 \\
Kick & 13 & Close & 32 & Hang & 10 \\
Lunge & 12 & & & Talk & 10 \\
Wave & 14  & & & Knock & 16 \\
& & & & Hole Punch &14\\
Total & 56 & Total & 64 & Total & 60\\
\hline
Total & 180 \\
\end{tabular}\\
\caption{ The number of instances of each event class in the dataset}
  \label{table:dataset}
\end{center}
\end{table}
\end{small}

\subsection{Analysis}

Three complete sets of annotations were obtained by three human annotators
using a GUI tool to mark the start and end frame of each instance of each
action.
Three annotators were used because events defined as state transitions are not
well defined as intervals, and so do not have clear start and end times.
All annotations and results are in the form of sets of tuples of (event label,
start frame, end frame), called intervals hereafter.
This allows comparison not only of machines against annotators, but annotators
against each other.
The disagreement among annotators can be seen in
Fig.~\ref{fig:fscore-comparison}.

Two intervals match if they overlap in time by more than some threshold.
This threshold is arbitrary, so all results are graphed as a function of
overlap threshold.
Treating one set of intervals as truth, and comparing another to it, the number
of true positives, false positives, and false negatives can be found.
From this the precision, recall, and F1 score can be computed.

\subsection{Comparison Methods}

We also ran on our dataset two recently published systems for which source code
is available: Action Bank \citep{SaCoCVPR2012} and C2 \citep{Kuehne11}.
These methods perform at state of the art levels on existing datasets.
Both of these systems are designed for 1 out of K classification on entire
videos, whereas the task at hand is K separate detection tasks over subsets of
long videos.
Therefore a simple extension to these methods was used.

The systems were trained using using video clips manually extracted from the
training videos, such that each clip depicted a single instance of an action.
An extra `no action' class was added to make it possible to recognize that
no event is recognized during an interval.
The test videos were split into a large number of short overlapping videos.
Each such video was classified by each of the two comparison methods.
Consecutive sets of matching labels were combined into a single interval.
No interval was created for sets of the `no action' label.

\subsection{Results}

\begin{figure}[t]
\centering
\subfigure[Hallway Scene]{
    \includegraphics[width=.54\textwidth]{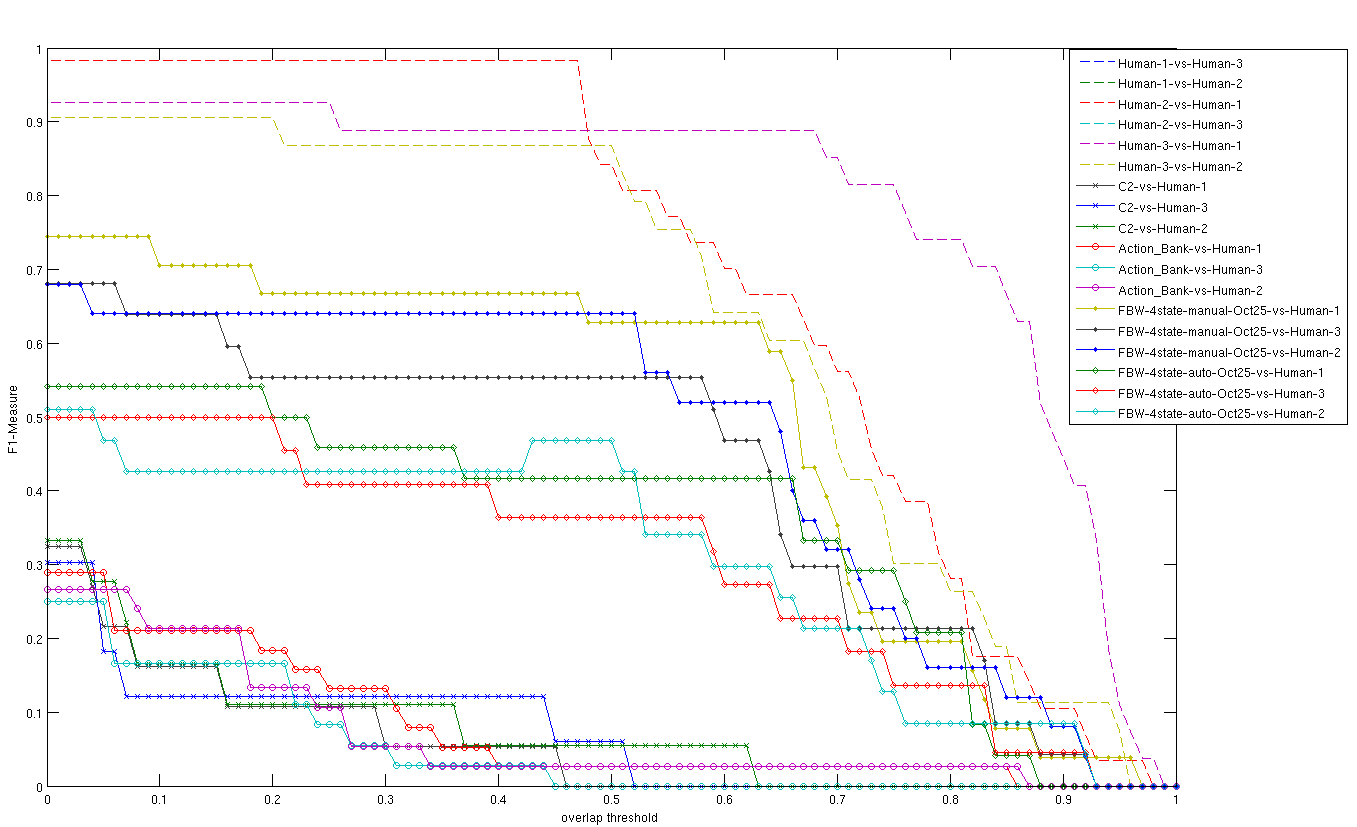}
  \label{fig:fscore-comparison-hallway}
}\\
\subfigure[Office Scene]{
    \includegraphics[width=.54\textwidth]{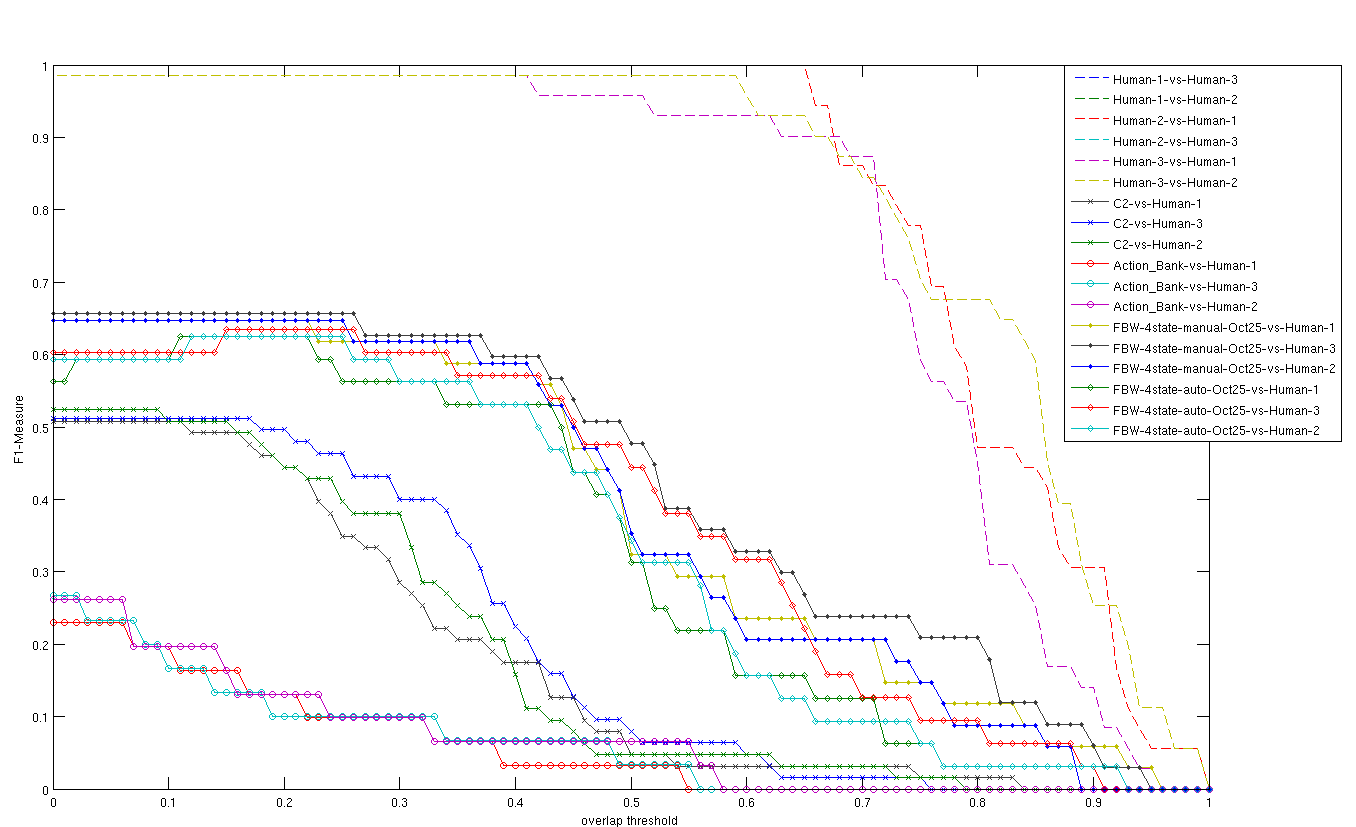}
  \label{fig:fscore-comparison-office}
}\\
\subfigure[Desk Scene]{
    \includegraphics[width=.54\textwidth]{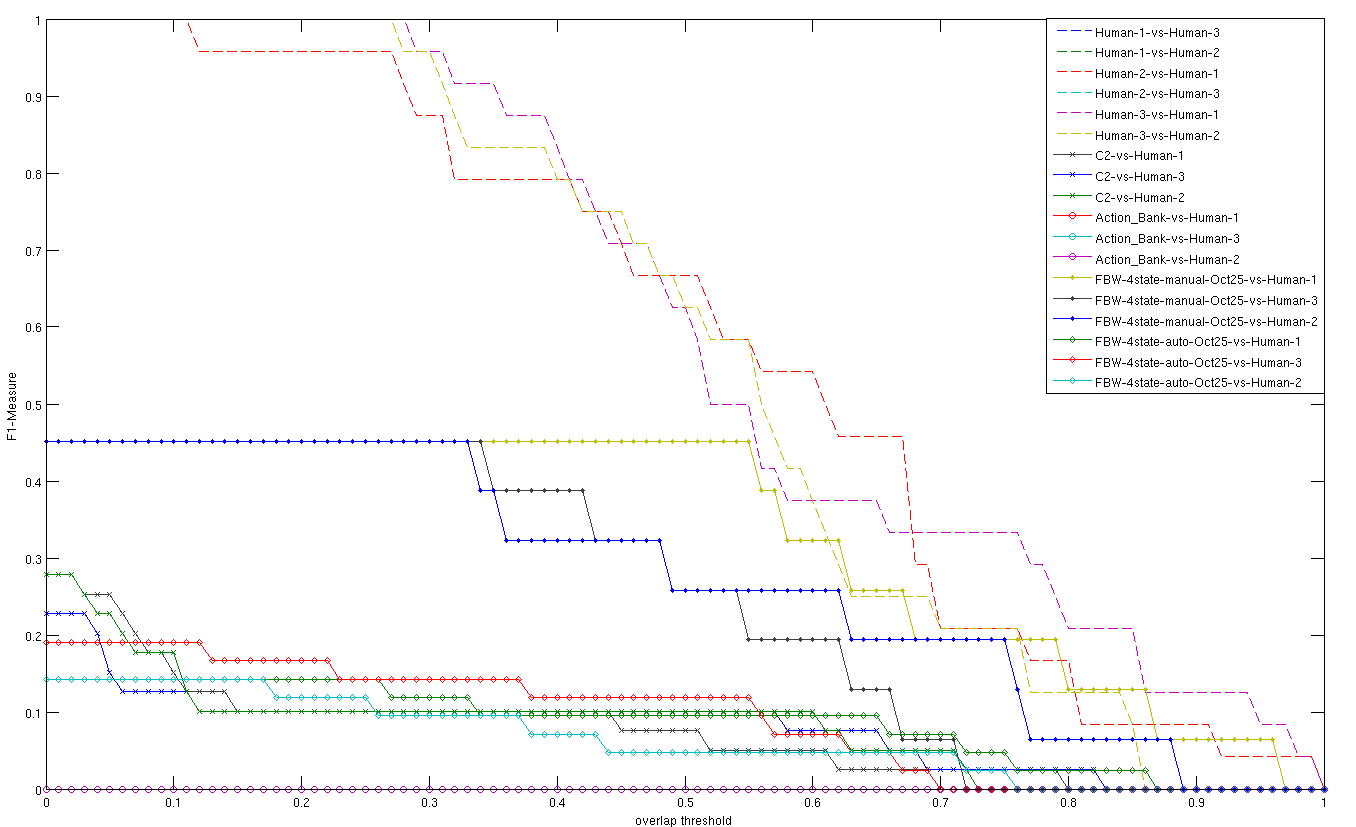}
  \label{fig:fscore-comparison-desk}
}\\
\subfigure[Entire Test Set]{
    \includegraphics[width=.54\textwidth]{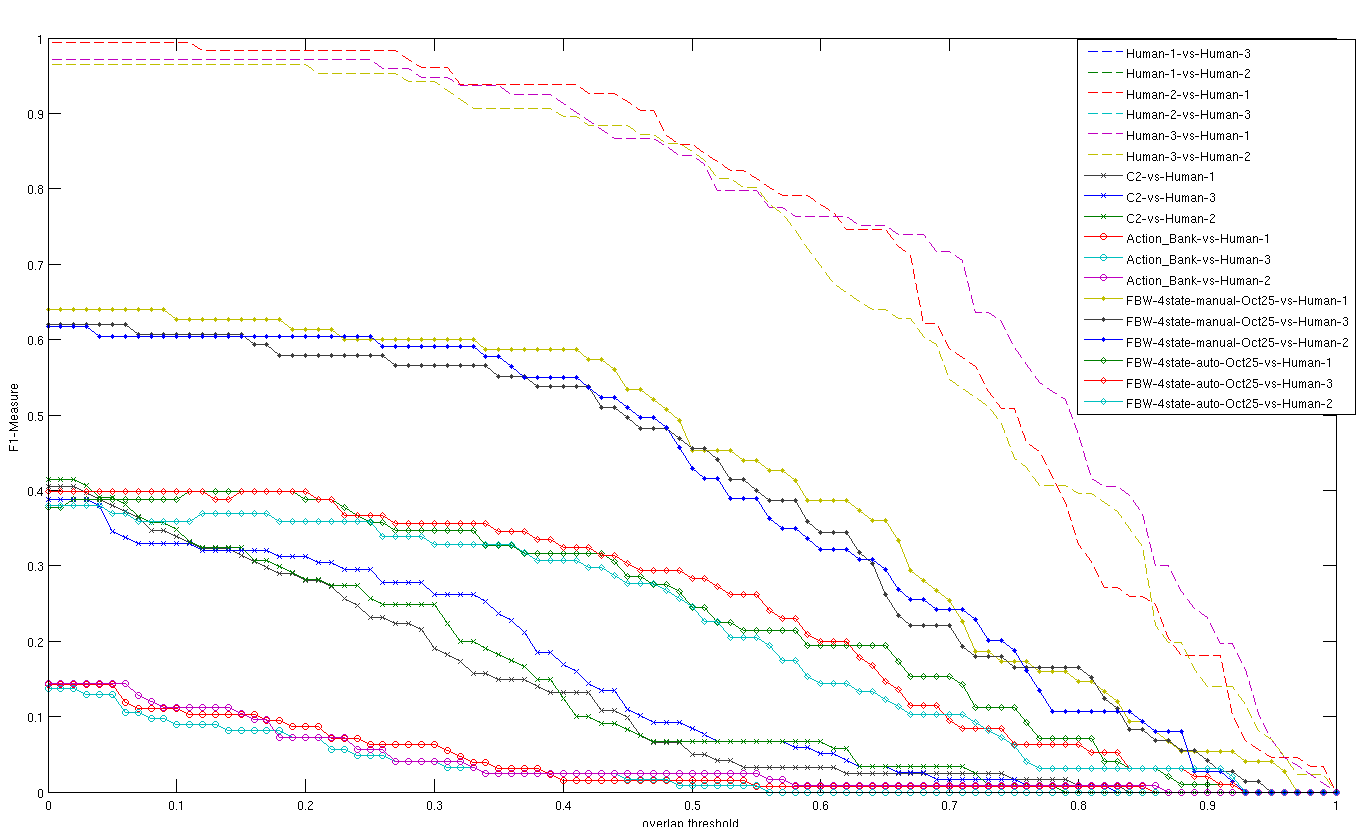}
  \label{fig:fscore-comparison-whole}
}
\caption{The F1-scores of human annotators, FBW with both manual and
    automatically generated tracks, C2 and Action Bank on each scene
    individually and the whole test test plotted as a function of overlap
    threshold.}
\label{fig:fscore-comparison}
\end{figure}

The proposed method outperformed both Action Bank and C2 both with automatic
and human annotated tracks.
This can be seen from Fig.~\ref{fig:fscore-comparison}, which plot the F1
measure of the proposed method both with manual and automatic tracks, C2,
Action Bank, and each of the three human annotators.
Note that the F1 measure is symmetric, so for example, the plot for Human-1
\vs\ Human-2 exactly overlaps the plot for Human-2 \vs\ Human-1, giving the
appearance that there are fewer plots than legend entries.
Each set of results was computed using each of the three human annotators as
the truth set.

\begin{figure}
  \begin{center}
    \includegraphics[width=.8\textwidth]{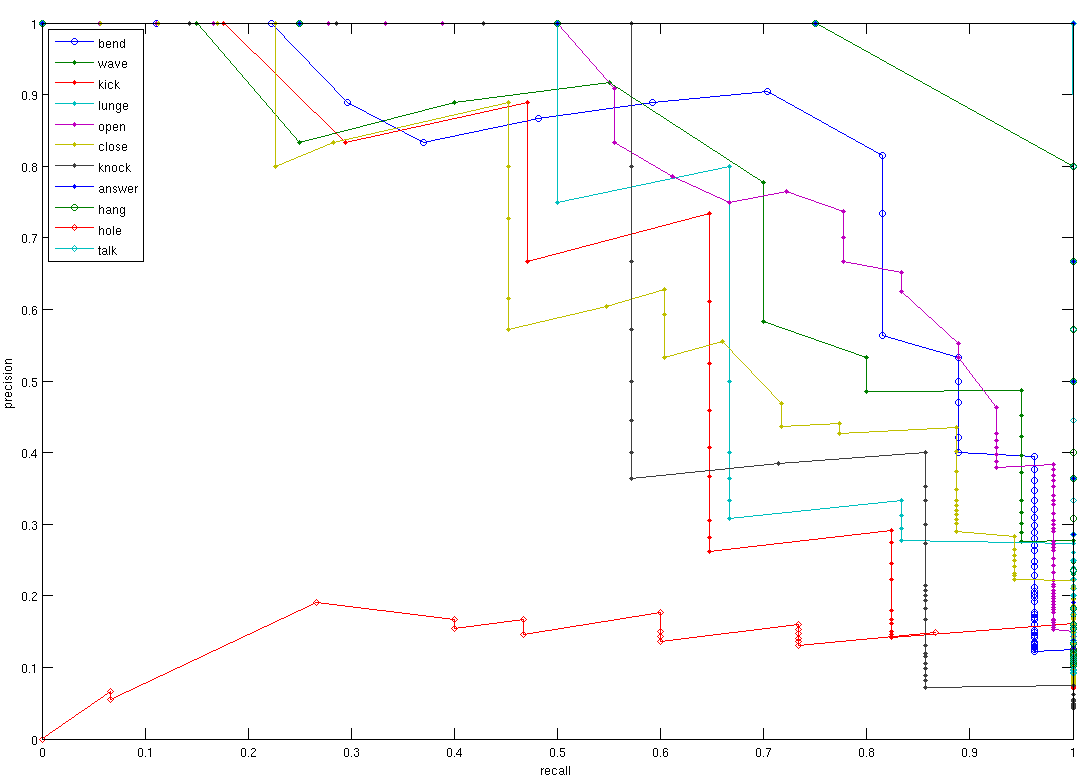}
  \end{center}

  \caption{Precision-recall graphs for each of the event models when using
    manual tracks.}
  \label{fig:precision-vs-recall}
\end{figure}

A precision-recall plot of the proposed method at a .1 overlap threshold with
manual tracks can be seen in Fig.~\ref{fig:precision-vs-recall}.\footnote{The
  video appendix includes a visualization of the performance of the system
  using the points from Fig.~\ref{fig:precision-vs-recall} which optimize the
  F1 score.}

\section{Conclusions and Future Work}

We have proposed a new method for event detection in video using HMMs with
object detectors as state output models.
We have also proposed a novel training procedure which allows object models for
each state to be learned during HMM training.
A new video corpus was filmed on which the performance of the proposed
method was compared against that of extensions to two recently published
methods, Action Bank and C2, as well as against the performance of human
annotators.
The proposed method outperformed both comparison methods on this dataset.

We are currently exploring the use of explicit duration modeling HMMs
\citep{Levinson:1986:CVD:1648817.1648829} to improve the HMMs' ability to model
the temporal sequences.
Modification by using multiple \citet{Felzenszwalb2010b,Felzenszwalb2010a} root
filters per state, or by allowing multiple branches in the HMM would allow each
branch to correspond to a family of viewing angles, improving the method's
ability to withstand viewpoint changes.

Another possible extension is the use of this method to characterize motion by
modifying the \citet{Felzenszwalb2010b,Felzenszwalb2010a} detector to use
histograms of optical flow vectors instead of histograms of gradients.
Testing the performance of this method with other object detectors is another
possible extension.

\ifnipsfinal
\subsubsection*{Acknowledgments}

This research was sponsored by the Army Research Laboratory and was
accomplished under Cooperative Agreement Number W911NF-10-2-0060.
The views and conclusions contained in this document are those of the authors
and should not be interpreted as representing the official policies, either
express or implied, of the Army Research Laboratory or the U.S. Government.
The U.S. Government is authorized to reproduce and distribute reprints for
Government purposes, notwithstanding any copyright notation herein.
\fi

\begin{small}
\bibliographystyle{abbrvnat}
\bibliography{arxiv2013c}
\end{small}
\end{document}